\def\year{2021}\relax
\documentclass[letterpaper]{article} 
\usepackage{aaai21}  
\usepackage{times}  
\usepackage{helvet} 
\usepackage{courier}  
\usepackage[hyphens]{url}  
\usepackage{graphicx} 
\usepackage{natbib}  
\usepackage{caption} 
\usepackage{comment}
\usepackage{amsmath,amssymb} 
\usepackage{microtype}
\usepackage{subfigure}
\usepackage{booktabs} 
\usepackage{multirow}
\usepackage{diagbox}
\usepackage{appendix}
\usepackage[linesnumbered, lined, boxed, commentsnumbered, ruled, vlined]{algorithm2e}
\usepackage[switch]{lineno} %

\newcommand{\eat}[1]{}

\newcommand{\smalltitle}[1]{ \vspace{1mm}{\noindent\textbf{#1.}\hspace{1mm}}}

\def\D{\mathcal{D}}
\def\M{\mathcal{M}}
\def\L{\mathcal{L}}

\usepackage{color}

\title{Split-and-Bridge: Adaptable Class Incremental Learning within a Single Neural Network}
\author{
    Jong-Yeong Kim, Dong-Wan Choi
    \\
}
\affiliations{
    Department of Computer Science and Engineering, Inha University, South Korea\\
    kjy93217@naver.com, dchoi@inha.ac.kr
}

\begin{document}

\maketitle

\begin{abstract}
Continual learning has been a major problem in the deep learning community, where the main challenge is how to effectively learn a series of newly arriving tasks without forgetting the knowledge of previous tasks. Initiated by \textit{Learning without Forgetting} (LwF), many of the existing works report that knowledge distillation is effective to preserve the previous knowledge, and hence they commonly use a soft label for the old task, namely a \textit{knowledge distillation (KD) loss}, together with a class label for the new task, namely a \textit{cross entropy (CE) loss}, to form a composite loss for a single neural network. However, this approach suffers from learning the knowledge by a CE loss as a KD loss often more strongly influences the objective function when they are in a competitive situation within a single network. This could be a critical problem particularly in a \textit{class incremental} scenario, where the knowledge across tasks as well as within the new task, both of which can only be acquired by a CE loss, is essentially learned due to the existence of a unified classifier. In this paper, we propose a novel continual learning method, called \textit{Split-and-Bridge}, which can successfully address the above problem by partially splitting a neural network into two partitions for training the new task separated from the old task and re-connecting them for learning the knowledge across tasks. In our thorough experimental analysis, our Split-and-Bridge method outperforms the state-of-the-art competitors in KD-based continual learning.
\end{abstract}
\section{Introduction} \label{sec:intro}

In recent years, deep neural networks (DNNs) have performed remarkably well in many practical applications like image or voice recognition \cite{HeZRS16},\cite{GravesMH13}, object detection \cite{HeGDG20}, and language translation \cite{SutskeverVL14}. A typical DNN learns the entire data for a fixed number of target tasks at once. However, in real-life applications encountering a dynamic stream of samples such as autonomous robots and unmanned vehicles, it is necessary to continuously incorporate a series of new tasks into the model being trained.

This problem is known as continual learning \cite{ParisiKPKW19}, which aims to incrementally train a model so that it can perform well on both previous tasks and new tasks. A major difficulty of continual learning in DNNs lies in the fact that the knowledge previously learned for old classes can severely be lost during the training process of new classes, often referred to as \textit{catastrophic forgetting} \cite{mccloskey1989catastrophic,mcclelland1995there}. In addition, such a loss of previous knowledge can be even more aggravated in \textit{class incremental learning} (CIL) where a single classifier should be incrementally unified. This is unlike \textit{task incremental learning} (TIL) where an inference is usually made with a \textit{task identity}, i.e., the prior knowledge of which task each sample belongs to, for as many classifiers as tasks that have been trained.

\begin{figure}[!t]
	\centering 
	\includegraphics[width=0.83\columnwidth]{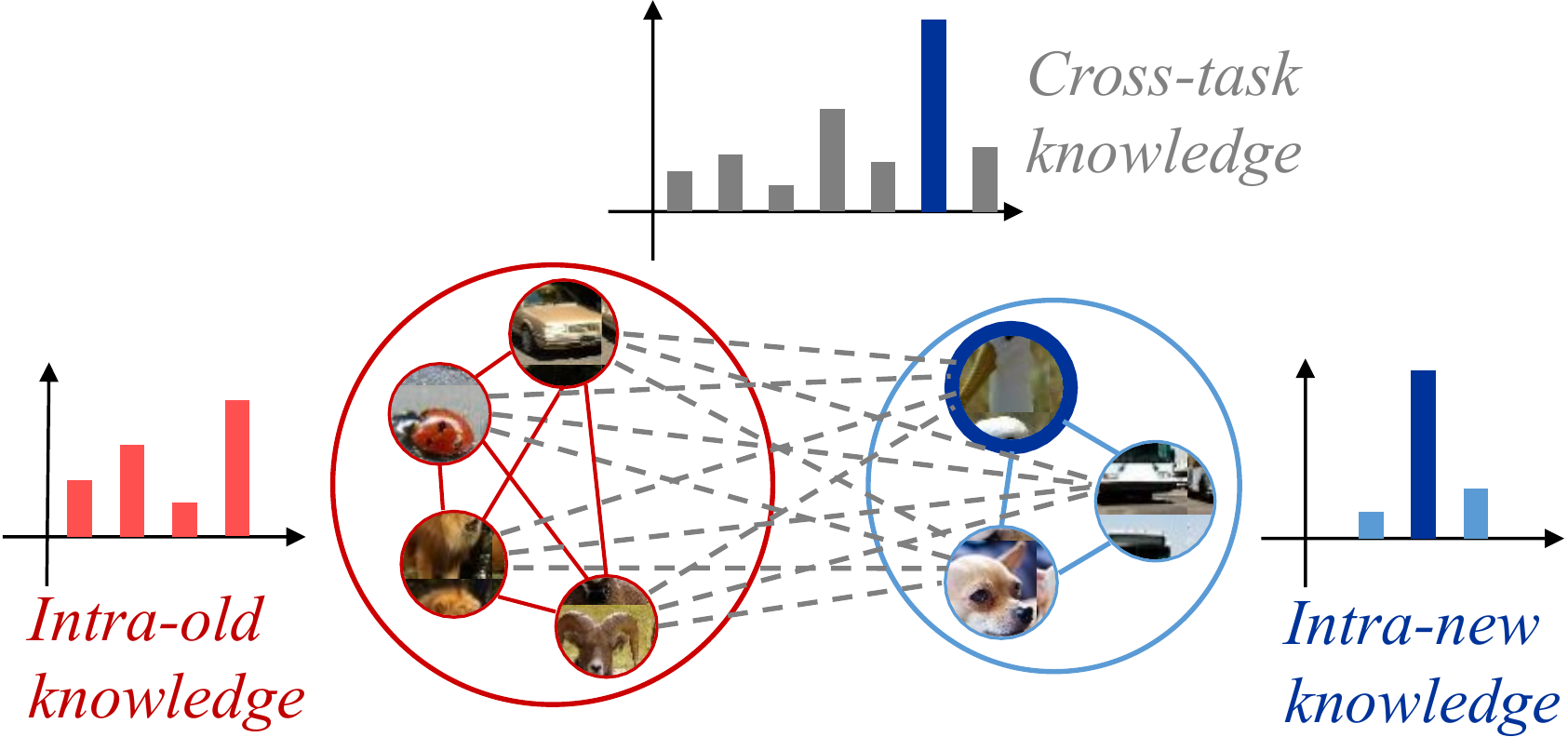}
	\caption{Three types of knowledge in CIL}
	\label{fig:knowledge}
\end{figure}

In general, we can think of CIL as the problem of learning three types of knowledge, namely \textit{intra-old}, \textit{intra-new}, and \textit{cross-task}, as shown in Figure \ref{fig:knowledge}. Either of the intra-old and the intra-new knowledge indicates how to discriminate classes only within the old task or the new task. On the other hand, the cross-task knowledge is about distinguishing every class in a particular task from the ones in the other task. Thus, we need all of the intra-old, intra-new, and cross-task knowledge for a unified classifier in CIL, but the cross-task knowledge is not necessary in TIL with a task identity provided at inference time. In this context, the goal of CIL is (i) to newly learn the intra-new knowledge as well as (ii) to incrementally update the cross-task knowledge (iii) without forgetting the previous intra-old knowledge.

\begin{figure}[t]
	\centering 
    \subfigure[\label{fig:framework:a}Standard KD-based CIL]{\hspace{10mm}\includegraphics[height=0.26\columnwidth]{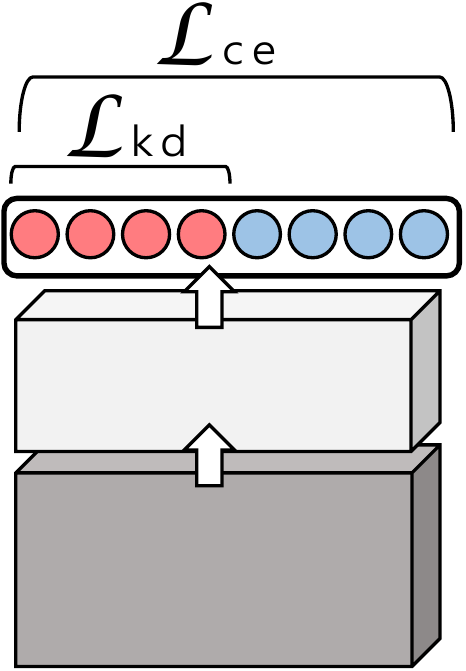}\hspace{10mm}}
    \subfigure[\label{fig:framework:b}Split-and-Bridge]{\hspace{5mm}\includegraphics[height=0.26\columnwidth]{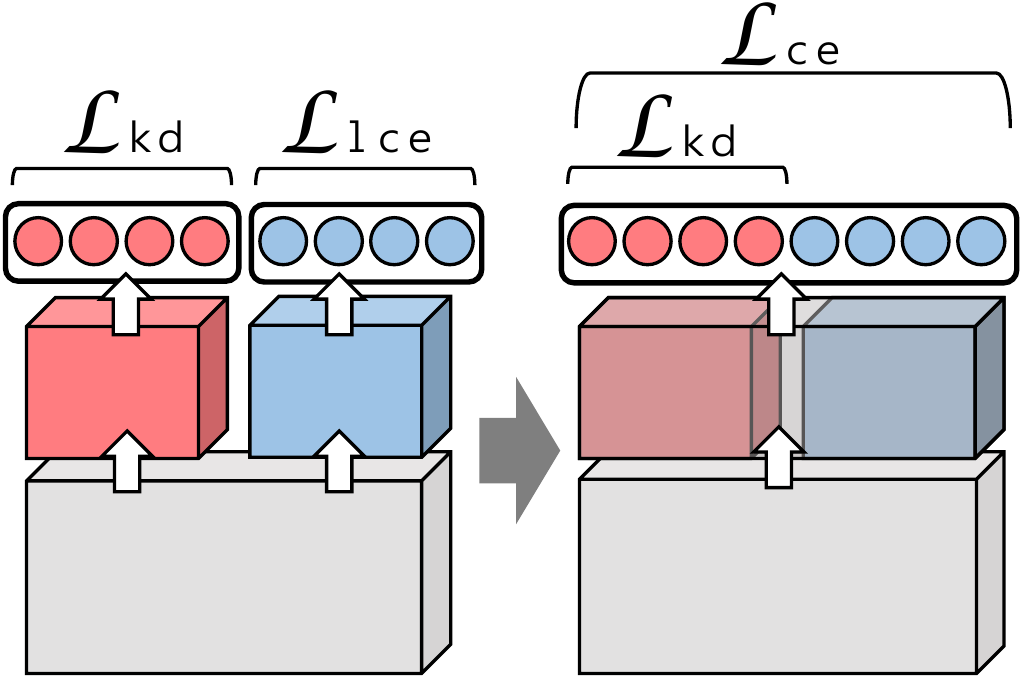}\hspace{5mm}} 
	\caption{Standard KD-based CIL vs. Split-and-Bridge}
	\label{fig:framework}
\end{figure}

As firstly introduced by \textit{Learning without Forgetting} (LwF) \cite{LiH16,LiH18a}, it has been reported that \textit{knowledge distillation} (KD) \cite{HintonVD15} is an effective strategy to preserve the intra-old knowledge. Also, the \textit{rehearsal} technique \cite{Lopez-PazR17,RebuffiKSL17} is often supplementary used to acquire the cross-task knowledge as well as to give an additional opportunity for learning the intra-old knowledge. The intra-new knowledge can obviously be learned by a given task-specific dataset. Many of the KD-based continual learning methods follow this standard training scheme shown in Figure \ref{fig:framework:a} even though they propose their own strategies to improve the way of making inference on the model already trained \cite{RebuffiKSL17,WuCWYLGF19,ZhaoXGZX20}. More specifically, for the intra-old knowledge, they mainly use a \textit{KD loss} with the soft label from the previous model along with a \textit{cross entropy (CE) loss} with the class label from stored exemplar samples, and learn the cross-task knowledge and the intra-new knowledge from a CE loss with the class labels from exemplar and new samples. 


Unfortunately, to be shown in our experiments, training a neural network is often more strongly influenced by a KD loss than by a CE loss.\eat{, when they are used at the same time.} This is intuitively because the KD loss is not intended to newly learn additional information but for the network to stay as it is. Also, in terms of the quantity of information, a soft label used by the KD loss carries a larger amount of information than a one-hot encoded class label in the CE loss. Consequently, the aforementioned KD-based method suffers from learning the intra-new and cross-task knowledge, both of which can only be acquired by a CE loss, due to the interference from a KD loss for the intra-old knowledge.

One natural solution for this problem would be to train only samples of new classes separately in another neural network, and then combine it into the model previously trained for old classes. To this end, we need to minimize two KD losses, each of which transfers the knowledge from either model being integrated. This way seems to effectively learn the intra-new knowledge without any intervention. At the same time, however, two KD losses for the integration would doubly disrupt learning  the cross-task knowledge from a CE loss, not to mention that we need such an extra model only temporarily used for new classes.

In this paper, we propose a novel CIL method, called \textit{Split-and-Bridge}, which can effectively learn all three types of knowledge in a way that learning each type of knowledge is much less interfered by learning the others even within a single neural network. As presented in Figure \ref{fig:framework:b}, Split-and-Bridge works in the following two phases, that is, split and bridge. In the split phase, we partially split a previously trained network into two partitions exclusive in upper layers yet sharing the same component in lower layers. In the mean time, for each exclusive partition, we separately learn either the intra-new knowledge or the intra-old knowledge without interrupting each other. Then, in the bridge phase, we re-connect (i.e., \textit{bridge}) those two partitions so that the cross-task knowledge can also be effectively learned without using two KD losses to integrate two pre-trained networks, i.e., the old model and new model.


In our experimental results, we show that Split-and-Bridge is superior to the state-of-the-art methods based on KD in CIL. In particular, our method turns out to be more adaptable to new classes than the existing methods, which is observed by the fact that the accuracy on the intra-new knowledge is fairly improved without loss of the knowledge previously learned.





\section{Related Work} \label{sec:related}

\smalltitle{Continual learning in DNNs}
Continual learning with DNNs has been mainly studied in the following two problem settings: \textit{class incremental learning} (CIL) and \textit{task incremental learning} (TIL). In CIL, a task identity should also be inferred for a unified classifier, whereas it is provided at test time in TIL. Both in CIL and TIL with DNNs, most of the existing works focus on how to overcome catastrophic forgetting, that is, the problem of preserving the previous knowledge when learning the new knowledge. The followings are the major branches of works on this problem strongly related to ours.

\smalltitle{Rehearsal methods} \textit{Rehearsal} methods \cite{Lopez-PazR17,RebuffiKSL17} store a subset of previous samples, and train them together with samples for a new task. This approach is known to be most effective to mitigate catastrophic forgetting in the CIL problem in which a single neural network needs to learn a sequence of tasks \cite{abs-1904-07734}, and therefore many state-of-the-art methods complementary leverage a rehearsal method and so does our Split-and-Bridge method.

\smalltitle{Parameter regularization} 
An alternative approach \cite{Kirkpatrick3521,ZenkePG17,ChaudhryDAT18,AljundiBERT18}  to address catastrophic forgetting is to regularize the parameters of a neural network so that more important parameters with respect to the previous task can be protected during training on each new task. The main drawback of this approach lies in the fact that it suffers from learning a long sequence of tasks as its main strategy is not to further change parameters from themselves that have been already trained. Consequently, it is reported that these methods do not work well especially in CIL, compared to the methods based on knowledge distillation (KD) \cite{abs-1904-07734}.

\smalltitle{Parameter isolation}
Mostly in a TIL scenario, we can also prevent catastrophic forgetting by separating model parameters for each task from the others. This idea motivates parameter isolation methods like \textit{PNN} \cite{RusuRDSKKPH16}, \textit{DEN} \cite{YoonYLH18}, \textit{PackNet} \cite{MallyaL18}, \textit{Piggyback} \cite{MallyaDL18_eccv}, \textit{HAT} \cite{SerraSMK18}, \textit{CGATE} \cite{AbatiTBCCB20}, all of which are designed to learn each task in an isolated part of the network and to make inference by using only the task-specific parameters selected by a task identity given at test time or inferred by an extra task classifier \cite{AbatiTBCCB20}. Another type of parameter isolation is to temporarily learn a new task in an extra network and then combine it into the previously learned model, which covers \textit{P\&C} \cite{Schwarz0LGTPH18}, \textit{DR} \cite{HouPLWL18}, \textit{DMC} \cite{ZhangZGLTHZK20}, and \textit{GD} \cite{LeeLSL19}. Our method is somewhat inspired by parameter isolation in the sense that we also temporarily split a network and then perform a unification process. However, none of the works do not deal with how to solve the CIL problem within a single neural network, which is the main goal of this work.

\smalltitle{KD-based methods} 
Similar to parameter regularization, KD-based methods basically aim to retain a pretrained model by transferring the previous knowledge distilled from the model. However, they differ from parameter regularization in that KD allows parameters to be well updated during training on a new task to find a better optima for both the previous and new task. Since this approach was firstly introduced by \textit{LwF} \cite{LiH16,LiH18a} for the TIL problem, there have been many variants following this standard LwF training scheme. \textit{iCaRL} \cite{RebuffiKSL17} first proposes a combination approach of rehearsal and KD. More recent approaches tend to focus on data imbalance problem between old classes and new classes in CIL, which includes \textit{WA} \cite{ZhaoXGZX20}, \textit{Bic} \cite{WuCWYLGF19}, \textit{LUCIR} \cite{HouPLWL19}, and \textit{EEIL} \cite{CastroMGSA18}. Although all these KD-based methods work effectively well to transfer the previous knowledge, we claim that these approaches could have undervalued the importance of the intra-new and cross-task knowledge, which should also be highly valued in the CIL problem we focus on.

\section{Preliminary} \label{sec:background}
This section first formally defines the class incremental learning (CIL) problem, and then describes the standard KD-based training method in the context of three essential types of knowledge in CIL as mentioned in the introduction.

\smalltitle{Class incremental learning problem}
We consider a sequence of tasks $T_1, T_2, ..., T_n$, where each $T_i$ at the $i$-th time step is a set of classes such that $T_i \cap T_j = \emptyset$ for any $i\neq j$, and carries its task-specific sample set $\D_i$. Each $\D_i$ consists of pairs $(\mathbf{x}, \mathbf{y})$ of input sample $\mathbf{x}$ and its one-hot encoded label $\mathbf{y}$. At any $t$-th step, we are given the new sample set $\D_t$ as well as an exemplar set $\M_t$ of a fixed size, which is a subset of previously observed samples, i.e., $\M_t \subseteq \D_1 \cup \D_2 \cup \cdots \cup \D_{t-1}$. Due to the limitation of memory space, it is usually assumed that  $|\M_t| \ll |\D_1 \cup \cdots \cup \D_{t-1}|$ in practice. Let $\Theta_i$ denote a neural network that have been trained from $T_1$ to $T_i$. Then, for each arrival of a new task $T_t$, the goal of the CIL problem is to newly train $\D_t$ with the help of $\M_t$ on the previously trained model $\Theta_{t-1}$ so that the resulting model $\Theta_{t}$ can work well with respect to all the classes in $T_1 \cup \cdots \cup T_t$.

As mentioned in the introduction, we can categorize the information required for each $\Theta_t$ into the three types of knowledge, namely \textit{intra-new}, \textit{intra-old}, and \textit{cross-task}. The intra-new knowledge is essential to identify the class of each new sample in $D_t$ from the other classes within $T_t$. Similarly, the intra-old knowledge represents how to discriminate samples within the set of old classes, i.e., $T_1 \cup \cdots \cup T_{t-1}$. On the other hand, the cross-task knowledge is necessary to distinguish samples across tasks, which informs how each new sample is different from old samples, and vice versa.

\smalltitle{Standard KD-based incremental learning}
In the literature, KD-based CIL methods mostly adopt the same strategy to learn each of the above three knowledge types. To illustrate, we first denote $o(\mathbf{x})$ as the output logit from a neural network for a given input sample $\mathbf{x}$. Then, we can define a \textit{cross entropy (CE) loss} on a model $\Theta$ for a given dataset $\D$ as:
\begin{equation} \label{eq:ce}
    \L_{ce}(\D, \Theta) = - \sum_{(\mathbf{x}, \mathbf{y}) \in \D}{\mathbf{y} \log p(\mathbf{x})},
\end{equation}
where $p(\mathbf{x})$ is the softmax probability vector of $o(\mathbf{x})$ such that $p(\mathbf{x})_i = \frac{e^{o(\mathbf{x})_i}}{\sum_j{e^{o(\mathbf{x})_j}}}$.

Similarly, we define a \textit{knowledge distillation (KD) loss} as follows:
\begin{equation}
    \L_{kd}(\D, \Theta) = - \sum_{(\mathbf{x}, \mathbf{y}) \in \D}{\hat{q}(\mathbf{x}) \log q(\mathbf{x})},
\end{equation}
where $q(\mathbf{x})$ is the softened probability of $o(\mathbf{x})$  such that $q(\mathbf{x})_i = \frac{e^{o(\mathbf{x})_i/\tau}}{\sum_j{e^{o(\mathbf{x})_j/\tau}}}$ for a temperature variable $\tau$, and $\hat{q}(\mathbf{x})$ indicates the soft label of $\mathbf{x}$ from the referenced model of $\Theta$, which is $\Theta_{t-1}$ at the $t$-th time step.

The standard KD-based method learns a new task $T_t$ by minimizing the following \textit{composite loss} function:
\begin{equation}
\lambda~\L_{kd}(\D_t \cup \M_t, \Theta_{t}) + (1-\lambda)~\L_{ce}(\D_t \cup \M_t, \Theta_{t}), \label{eq:stdloss}
\end{equation}
where $\lambda$ is a hyperparameter balancing between two losses, and usually set to $\frac{C_{old}}{C_{old}+C_{new}}$ such that $C_{old} = |T_1 \cup \cdots \cup T_{t-1}|$ and $C_{new} = |T_t|$. We can further identify which part of this loss function is utilized to acquire each type of knowledge, and our understanding is as follows:
\begin{itemize}
    \item Intra-old knowledge: $\L_{kd}(\D_t \cup \M_t, \Theta_{t})$ + $\L_{ce}(\M_t, \Theta_{t})$
    \item Intra-new knowledge: $\L_{ce}(\D_t, \Theta_{t})$
    \item Cross-task knowledge: $\L_{ce}(\D_t \cup \M_t, \Theta_{t})$
\end{itemize}
Thus, both the intra-new and the cross-task knowledge have to be learned by a CE loss, but this could not be very effective as a KD loss for the intra-old knowledge is likely to dominate the final loss function. This is a reasonable conjecture in that the soft label $\hat{q}(\mathbf{x})$ will carry a larger amount of information than the one-hot encoded label $\mathbf{y}$, leading to more updates by $\L_{kd}$ than by $\L_{ce}$.

\begin{table}[t]
\begin{center}
\small 
    \begin{tabular}{c||c|c|c|c|c}
        \hline
        \multirow{3}{*}{Loss} & \multicolumn{5}{c}{Accuracy (\%)} \\
        \cline{2-6}
         &  \multirow{2}{*}{\textbf{Overall}} & \multirow{2}{*}{\textbf{Old}} & \multirow{2}{*}{\textbf{New}} & \textbf{Intra-} & \textbf{Intra-}\\
         &  &  &  & \textbf{old} & \textbf{new}\\
        \hline
        $\L_{ce}$ & 74.25 &  58.9 & \textbf{89.6} & 82.10 & \textbf{90.85}\\
        $\L_{kd}$ + $\L_{ce}$  & \textbf{75.67} & \textbf{65.6} & 85.75 & \textbf{84.85} & 88.30\\
        \hline
    \end{tabular}
    \caption{Four types of accuracy after incrementally learning two tasks, each consisting of 20 classes, on ResNet-18 with CIFAR-100, where the accuracy for the first task is 85.05\%}
    \label{tab:stdloss}
\end{center}
\end{table}

To examine our conjecture, we conduct a simple experiment on learning CIFAR-100 \cite{krizhevsky2009learning} on ResNet-18 \cite{HeZRS16} with two incremental tasks, each consisting of 20 classes. Then, we compare the resulting accuracy of using $\L_{kd}$ + $\L_{ce}$ to that of using only $\L_{ce}$, where the accuracy is further divided into four categories; old, new, intra-old, and intra-new, to show how well each type of knowledge is learned by each loss function. When measuring either the intra-old or intra-new accuracy, we locally compare only the output probabilities corresponding to either the first (i.e., old) task or the second (i.e., new) task. As shown in Table \ref{tab:stdloss}, it is well observed that the new accuracy and the intra-new accuracy of using $\L_{kd}$ + $\L_{ce}$ gets lower than those of using only $\L_{ce}$. Thus, even though the standard KD-based method (i.e., $\L_{kd}$ + $\L_{ce}$) improves the overall accuracy partly because of the preservation of the intra-old knowledge, it is achieved at the sacrifice of the adaptability to the new task. 
\section{Proposed Adaptable Incremental Learning} \label{sec:ours}
In this section, we propose our \textit{Split-and-Bridge} method for the CIL problem, which can effectively learn all three types of knowledge.

\smalltitle{Motivation}
As described in the preliminary section, the standard KD-based method suffers from learning the new knowledge by a CE loss due to the interference from the KD loss. With respect to the intra-new knowledge, a possible solution would be to separately learn a new task with an extra network, and then integrate its intra-new knowledge with the previous knowledge by distilling from two teacher models (i.e., the old model and the new model). However, at this time, the cross-task knowledge to be learned by another CE loss is doubly disturbed by these two KD losses. 

This motivates us to propose a two phase learning method within a single network, where we first (i) \textit{partially split} the network into two partitions for a separated learning, namely \textit{split phase}, and then (ii) \textit{re-connect} these trained partitions to additionally learn the cross-task knowledge, namely \textit{bridge phase}. By doing so, we do not only learn the intra-task knowledge in an isolated partition without any competition between losses, but also acquire the cross-task knowledge without having to use double KD losses.

\begin{figure}[t]
	\centering 
	\includegraphics[width=0.88\columnwidth]{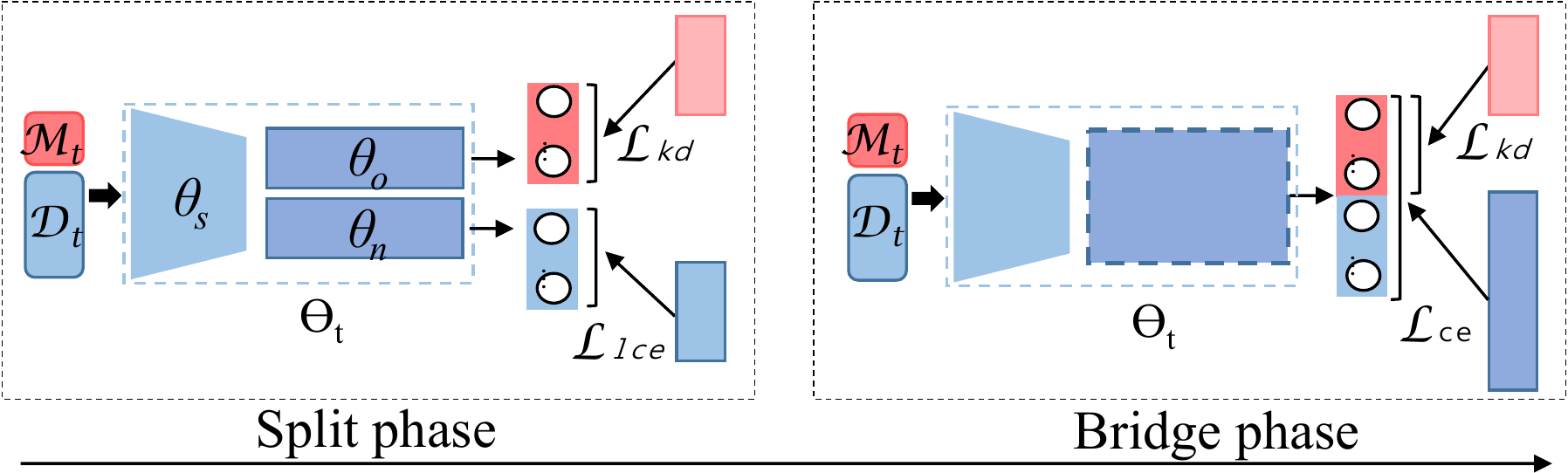}
	\caption{Two-phase learning of Split-and-Bridge}
	\label{fig:splitnet}
\end{figure}

\subsection{Split Phase}
\smalltitle{Separated learning within a single network}
The goal of \textit{split phase} is to learn the intra-new knowledge as independently as possible from the task of preserving the intra-old knowledge without using an extra neural network. To this end, we need to re-organize the given network $\Theta_t$ to have two disjoint branches at upper layers, denoted by $\theta_o$ and $\theta_n$, coming from a shared common feature extractor spanning lower layers, denoted by $\theta_s$, as shown in Figure \ref{fig:splitnet}. Once $\Theta_t$ is successfully transformed into the branched network, denoted by $\langle \theta_s, [\theta_o, \theta_n]\rangle_t$, we can learn both the intra-new and intra-old knowledge by the following loss function:
\begin{equation} \label{eq:kdlce}
\L_{kd}(\D_t \cup \M_t, \langle\theta_{s},\theta_{o}\rangle_t) + \L_{lce}(\D_t, \langle\theta_{s},\theta_{n}\rangle_t).
\end{equation}
Through $\L_{kd}$, we distill the intra-old knowledge from the previous model, i.e., $\Theta_{t-1}$, into a part of the model consisting of the shared component followed by the disjoint partition for the old task, i.e., $\langle\theta_{s},\theta_{o}\rangle_t$. Separately, the intra-new knowledge is learned on the other part of the model consisting of the shared component and the other partition, i.e., $\langle\theta_{s},\theta_{n}\rangle_t$. Thus, with respect to $\theta_n$ and $\theta_o$, each learning is completely independent, while both of learning processes are performed on $\theta_s$. It is reasonable to have this shared component as the features at lower levels are usually applicable to every task and therefore such common features can effectively help to learn either task as well.

\begin{figure}[t]
	\centering 
	\includegraphics[width=0.92\columnwidth]{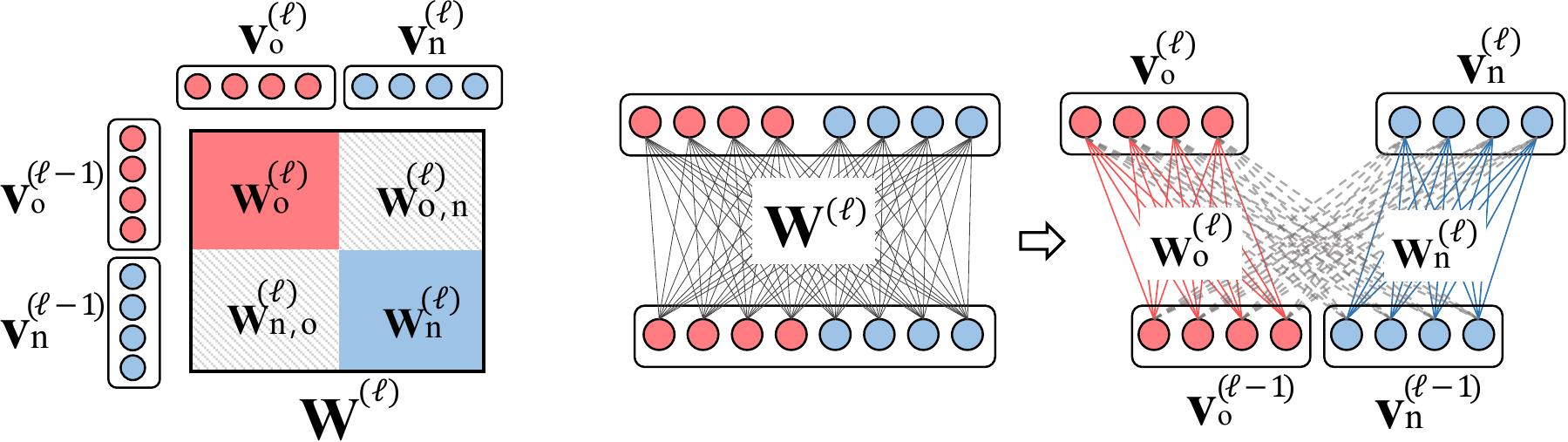}
	\caption{Sparsification across tasks}
	\label{fig:sparsification}
\end{figure}

Note that we use a different type of loss, called \textit{localized cross entropy (LCE)} and denoted by $\L_{lce}$, for learning the intra-new knowledge in this branched network. This loss is similar to the normal CE loss, but it only takes into account the local probabilities within the new task in order to focus on the intra-new knowledge as defined:
\begin{equation}
    \L_{lce}(\D_t, \Theta) = - \sum_{(\mathbf{x}, \mathbf{y}) \in \D_t}{\mathbf{y_t} \log p_t(\mathbf{x})},
\end{equation}
where $p_t(\mathbf{x})$ represents a softmax output locally computed using only the sub-logit corresponding to the task $T_t$, and $\mathbf{y}_t$ is similarly a sliced hard label corresponding to $T_t$. Recall that $p(\mathbf{x})$ and $\mathbf{y}$ are defined for the all the classes in Eq. (\ref{eq:ce}), but $p_t(\mathbf{x})$ and $\mathbf{y_t}$ are differently defined as $p_t(\mathbf{x})_i = \frac{e^{o(\mathbf{x})_i}}{\sum_j{e^{o(\mathbf{x})_j}}}~~\textit{and}~~\mathbf{y_t}_i = \mathbf{y}_i$ such that $i, j \in [C_{old} + 1, C_{old} + C_{new}]$, where $C_{old}$ is the number of old classes and $C_{new}$ is the number of new classes.



\smalltitle{Weight sparsification across tasks}
Then, how do we get a given network $\Theta_t$ to have a partially separated structure $\langle \theta_s, [\theta_o, \theta_n]\rangle_t$? 
In a simple way, we can randomly select two disjoint partitions and disconnect all the weights between them. However, such a simple method can cause a considerable loss of the previously trained knowledge. Our solution is instead to make those weights across partitions as sparse as possible during learning the intra-new knowledge as well as distilling the intra-old knowledge. This idea is inspired by \textit{SplitNet} \cite{KimPKH17}, which is originally introduced to build a tree-structured network with multiple branches such that similar classes are assigned to the same part of the network parameters. In the CIL problem setting, our goal is to prevent mutual interference between tasks during learning a new task, instead of grouping similar classes.

As illustrated in Figure \ref{fig:sparsification}, let us first consider a weight matrix $\mathbf{W}^{(\ell)}$ in layer $\ell$, which takes the output $\mathbf{v}^{(\ell-1)}$ of the $(\ell-1)$-th layer as its input vector and produces its output $\mathbf{v}^{(\ell)}$ to be the input of the $(\ell+1)$-th layer. Also, let $S$ denote the index of the last layer covered by $\theta_s$, and then we have to split each $\mathbf{W}^{(\ell)}$ into two partitions for each layer $\ell \in [S+1, L]$, where $L$ indicates the final layer of the network. Splitting each $\mathbf{W}^{(\ell)}$ means that we divide its input nodes $\mathbf{v}^{(\ell-1)}$ and output nodes $\mathbf{v}^{(\ell)}$ into two disjoint groups with a particular ratio (e.g., 1:1), namely $\mathbf{v}_o^{(\ell-1)}$ and $\mathbf{v}_o^{(\ell)}$ for the old task and $\mathbf{v}_n^{(\ell-1)}$ and $\mathbf{v}_n^{(\ell)}$ for the new task, and then disconnect all the weights between either $\mathbf{v}_o^{(\ell-1)}$ and $\mathbf{v}_n^{(\ell)}$ or $\mathbf{v}_n^{(\ell-1)}$ and $\mathbf{v}_o^{(\ell)}$ as shown in Figure \ref{fig:sparsification}. As mentioned above, to preserve the previous knowledge as much as possible in such a splitting process, we train a single network $\Theta_t$ by the following loss function, which can make weights to be disconnected as sparse as possible, while simultaneously trying to put each of the intra-old and the intra-new knowledge into two separated partitions:
\begin{eqnarray} \label{eq:splitloss}
\L_{kd}(\D_t \cup \M_t, \Theta_t) + \L_{lce}(\D_t, \Theta_t) \nonumber \\
+~~\gamma\sum_{\ell=S+1}^{L}(||\mathbf{W}^{(\ell)}_{o,n}||_2 + ||\mathbf{W}^{(\ell)}_{n,o}||_2),
\end{eqnarray}
where $\mathbf{W}^{(\ell)}_{o,n}$ and $\mathbf{W}^{(\ell)}_{n,o}$ indicate two sub-matrices whose weights should be disconnected, as defined:
\begin{eqnarray}
\mathbf{W}^{(\ell)}_{o,n}  & = &  \{w_{ij} \in \mathbf{W}^{(\ell)} |~i \in \mathbf{v}_o^{(\ell-1)} \wedge j \in \mathbf{v}_n^{(\ell)}\}  \nonumber \\
\mathbf{W}^{(\ell)}_{n,o}  & = & \{w_{ij} \in \mathbf{W}^{(\ell)} |~i \in \mathbf{v}_n^{(\ell-1)} \wedge j \in \mathbf{v}_o^{(\ell)}\}. \nonumber
\end{eqnarray}
By Eq. (\ref{eq:splitloss}), $\Theta_t$ is jointly optimized by $\L_{kd}$, $\L_{lce}$, and a regularization term for sparsification. Note that the regularization term imposes $l^{2}$ norm on two sub-matrices to be disconnected, and therefore minimizing this loss makes those weights as small as possible. $\gamma$ controls the strength of this sparse regularization. Once the training of weight sparsification is performed enough, we explicitly disconnect all $\mathbf{W}^{(\ell)}$'s into two partitions $\mathbf{W}_o^{(\ell)}$'s and $\mathbf{W}_n^{(\ell)}$'s, all of which together constitute $\theta_o$ and $\theta_n$, respectively, for further training on the resulting branched network by minimizing Eq. (\ref{eq:kdlce}).

Note that this sparsification process can also help to prevent overfitting in earlier incremental steps. If we want our model to accept as many tasks as possible, it is somewhat unavoidable to start with a high-capacity model. However, such a large model tends to overfit by memorizing patterns of samples belonging to a few classes in earlier steps. Our regularization term for sparsification can mitigate this overfitting problem particularly when we train initial tasks consisting of only a few classes  on a model with high capacity.

\begin{algorithm}[htb]
   \caption{Split-and-Bridge Incremental Learning}
   \label{alg:sbtrain}
   $\Theta_0 \leftarrow$ a neural network randomly initialized\;
   $\M_1 \leftarrow \emptyset$\;
   \ForEach{incremental task $T_t$}
   {
    \KwIn{model $\Theta_{t-1}$, the task-specific data $\D_t$}
    \KwOut{model $\Theta_{t}$ }
        $\Theta_{t} \leftarrow \Theta_{t-1}$\;

        \If{$t = 1$}
        {
            Train $\Theta_{t}$ by minimizing $\L_{ce}(\D_t, \Theta_{t})$\;
        } 
        \Else{
            Train $\Theta_{t}$ by minimizing Eq. (\ref{eq:splitloss})\;
            Explicitly disconnect $\Theta_t$ into $\langle\theta_s, [\theta_o, \theta_n]\rangle_t$\;
            Train $\langle\theta_s, [\theta_o, \theta_n]\rangle_t$ by minimizing Eq. (\ref{eq:kdlce})\;
            Re-connect $\theta_o$ and $\theta_n$ to form $\Theta_t$\;
            Train $\Theta_t$ by minimizing Eq. (\ref{eq:stdloss})\;
        }
        $\M_{t+1} \leftarrow$ random sample from $\M_{t} \cup \D_t$\;
   }
\end{algorithm}

\smalltitle{Adaptive split ratio}
Now, the question is how much we allocate the partition of each task. In the final layer (i.e., when $\ell = L$), there is no choice but to assign the final outputs of old classes to $\mathbf{v}_o^{(L)}$ and set $\mathbf{v}_n^{(L)}$ to those of new classes. Other than the final layer, we propose an \textit{adaptive splitting scheme} such that $|\mathbf{v}^{(\ell)}_{o}| : |\mathbf{v}^{(\ell)}_{n}|$ = $\rho~C_{old} :(1-\rho)~C_{old} + C_{new}$, where $\rho$ is a hyperparameter that controls the allocation rate for the old task, which depends on models, total number of steps, etc. Once $|\mathbf{v}^{(\ell)}_{n}| < 1$, we consider $\ell$ to be a layer shared by the intra-old and intra-new knowledge. In our experiments, we set $\rho$ to a value between $1.0$ and $1.4$.

\subsection{Bridge Phase}
Given $\langle \theta_s, [\theta_o, \theta_n]\rangle_t$ that has separately learned the intra-old and intra-new knowledge, we re-connect two partitions $\theta_o$ and $\theta_n$ in order to learn the cross-task knowledge between them in the bridge phase. To this end, we first initialize all the weights that have been removed in the split phase to be zero. This is intuitively because a random initialization can yield erroneous cross-task information and this would happen to break the model $\langle \theta_s, [\theta_o, \theta_n]\rangle_t$ trained well in the split phase. Our intention in the bridge phase is to learn any new information across tasks on these zero-initialized \textit{bridge} weights if necessary. 

In order to train this re-connected network, we minimize the same loss function of Eq. (\ref{eq:stdloss}) as the standard KD-based method. At this time, however, we train the model that has already learned the intra-new knowledge as well as the intra-old knowledge in the split phase, not a model trained only for the old task. Thus, since the intra-new knowledge is already there in the target model being trained, learning by Eq. (\ref{eq:stdloss}) is like an auxiliary training process as for the intra-new knowledge.

In addition, the KD loss now transfers not only the intra-old knowledge but also the \textit{common knowledge} between old and new classes as its referenced model is now $\langle \theta_s, \theta_o \rangle_t$ containing the shared component $\theta_s$, not $\Theta_{t-1}$ owning only the intra-old knowledge. As a result, the KD loss would even help to learn the cross-task knowledge, which is mainly learned by the overall CE loss.


\smalltitle{Overall algorithm} 
In summary, Algorithm \ref{alg:sbtrain} outlines how Split-and-Bridge trains a given neural network for each incremental task. Learning the first task is typical training by a CE loss (Lines 5-6). After that, we first train $\Theta_t$ to have a partially split network (Lines 8-9), and then perform a separated learning on it (Line 10). Finally, we re-connect the split partitions for further learning the cross-task knowledge (Lines 11-12). It is noteworthy that the KD loss in Lines 8 and 10 uses $\Theta_{t-1}$ as its reference model, but the reference model is changed to $\langle \theta_s, \theta_o \rangle_t$ for Line 12 once the branched network is successfully trained.

\section{Experiments} \label{sec:exp}

\begin{figure*}[t]
	\centering
	{
    \includegraphics[height=2.5mm]{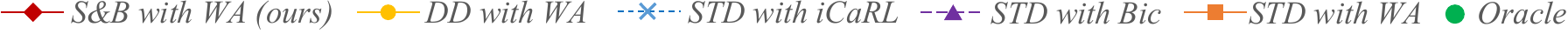}	\vspace{1mm}

		\begin{tabular}{cccc}
			\subfigure[\label{fig:cifar:a}2 tasks] {\includegraphics[height=32mm]{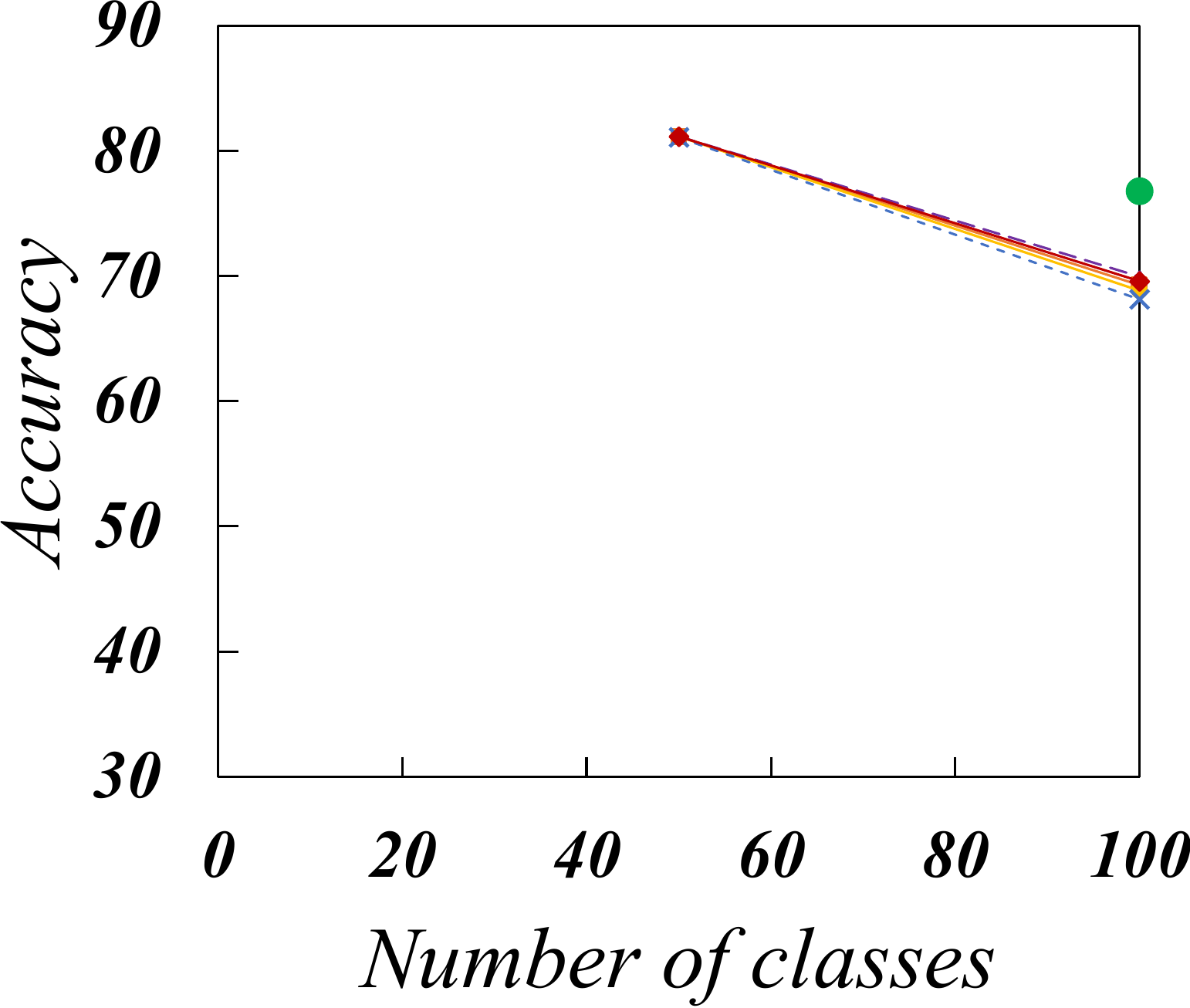}}  & 
			\subfigure[\label{fig:cifar:b}5 tasks] {\includegraphics[height=32mm]{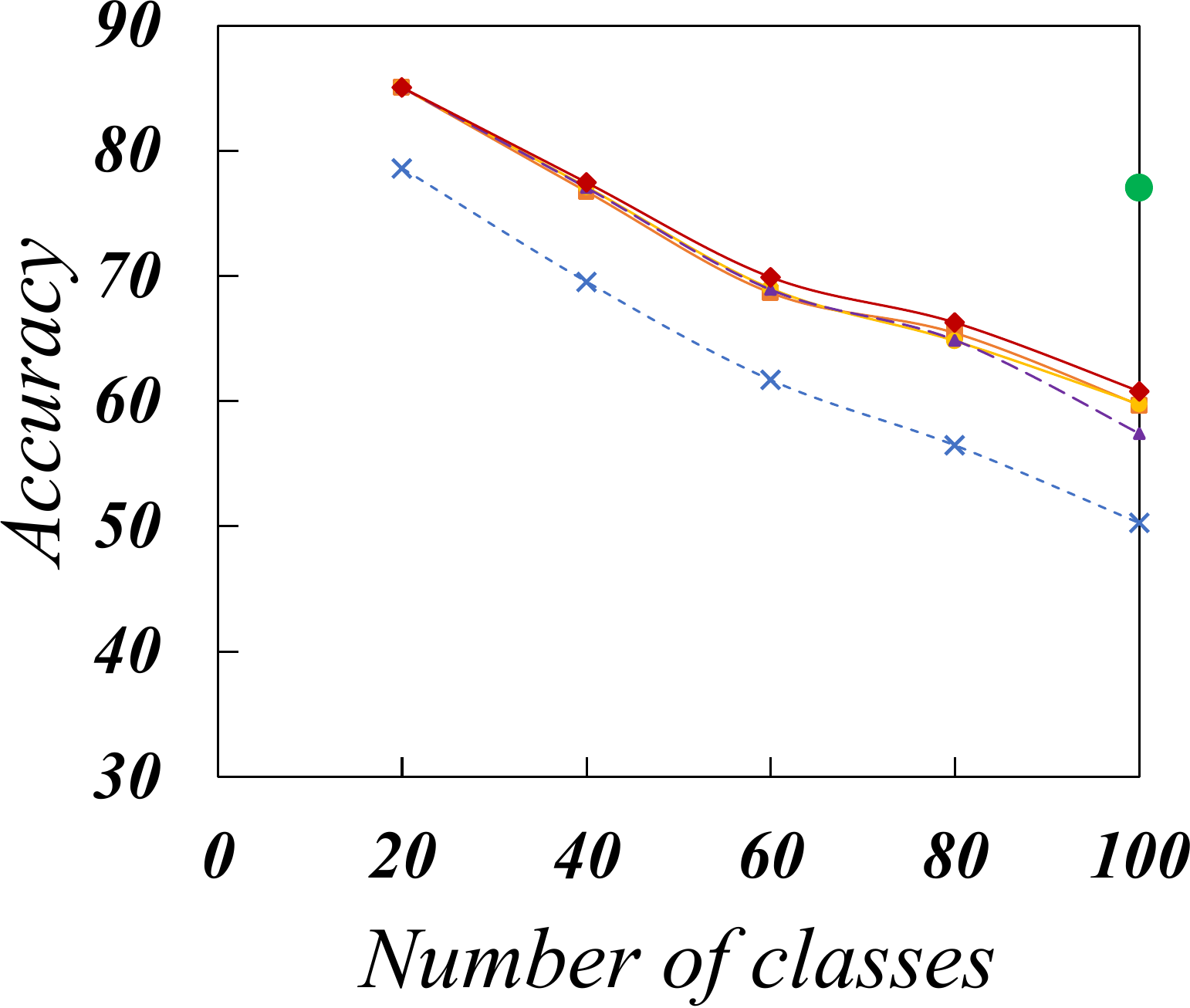}}   &
			\subfigure[\label{fig:cifar:c}10 tasks] {\includegraphics[height=32mm]{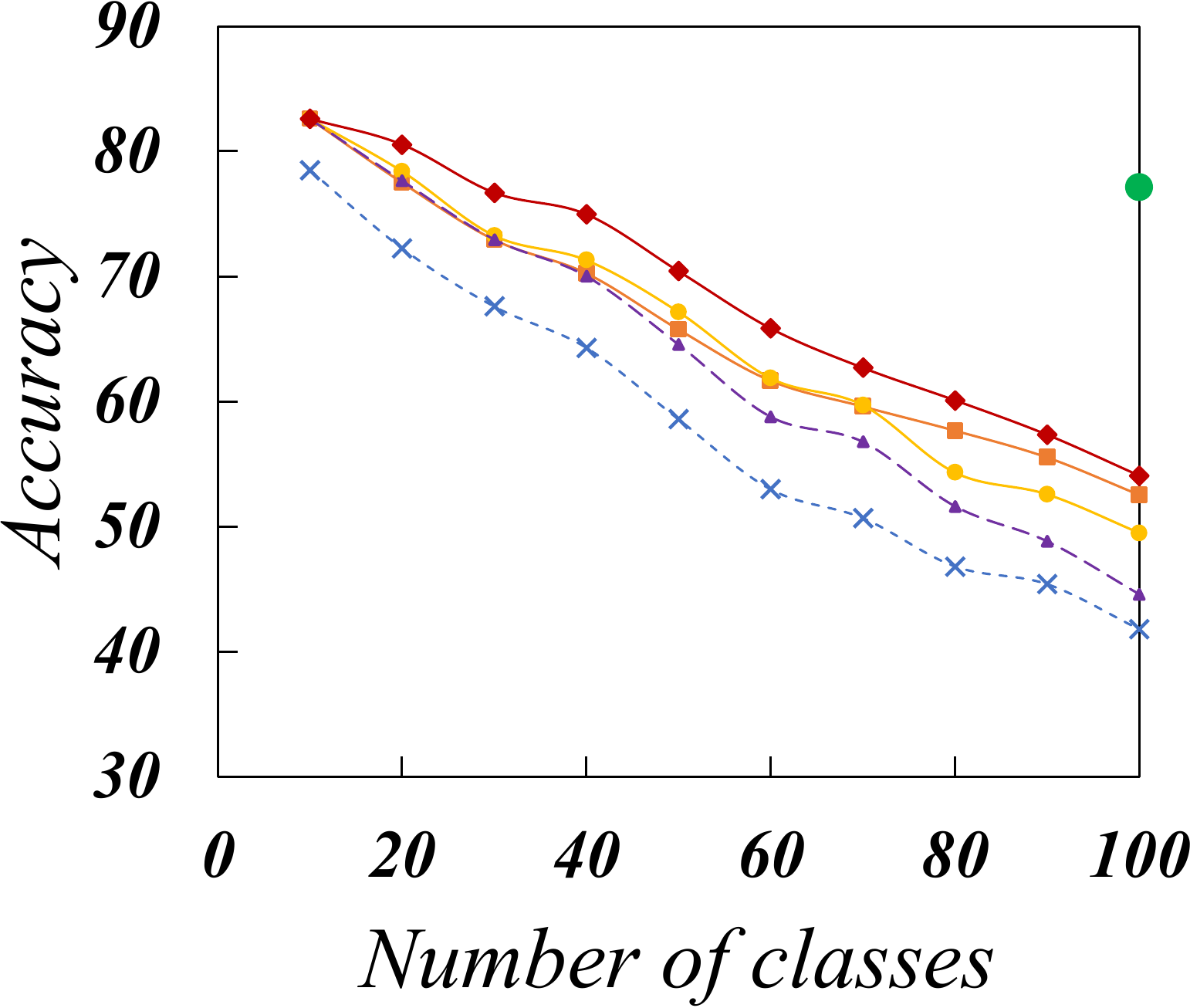}}  &
			\subfigure[\label{fig:cifar:d}20 tasks] {\includegraphics[height=32mm]{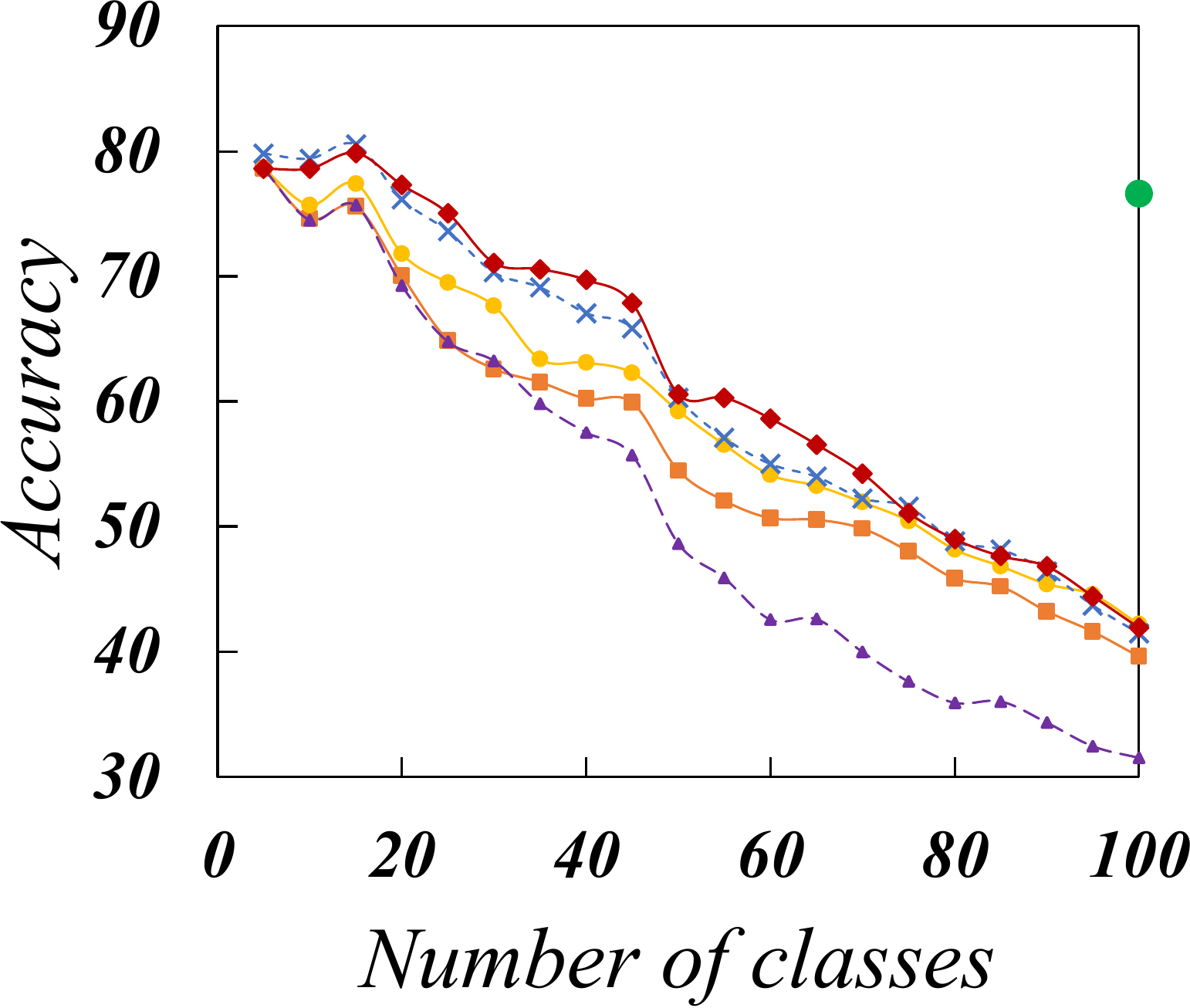}}  
		\end{tabular}
		\vspace{-3mm}
		\caption{Comparison on the accuracy for each incremental task using CIFAR-100}
		\label{fig:cifar}
	}
	\centering
	{
		\begin{tabular}{cccc}
			\subfigure[\label{fig:tiny:a}2 tasks] {\includegraphics[height=32mm]{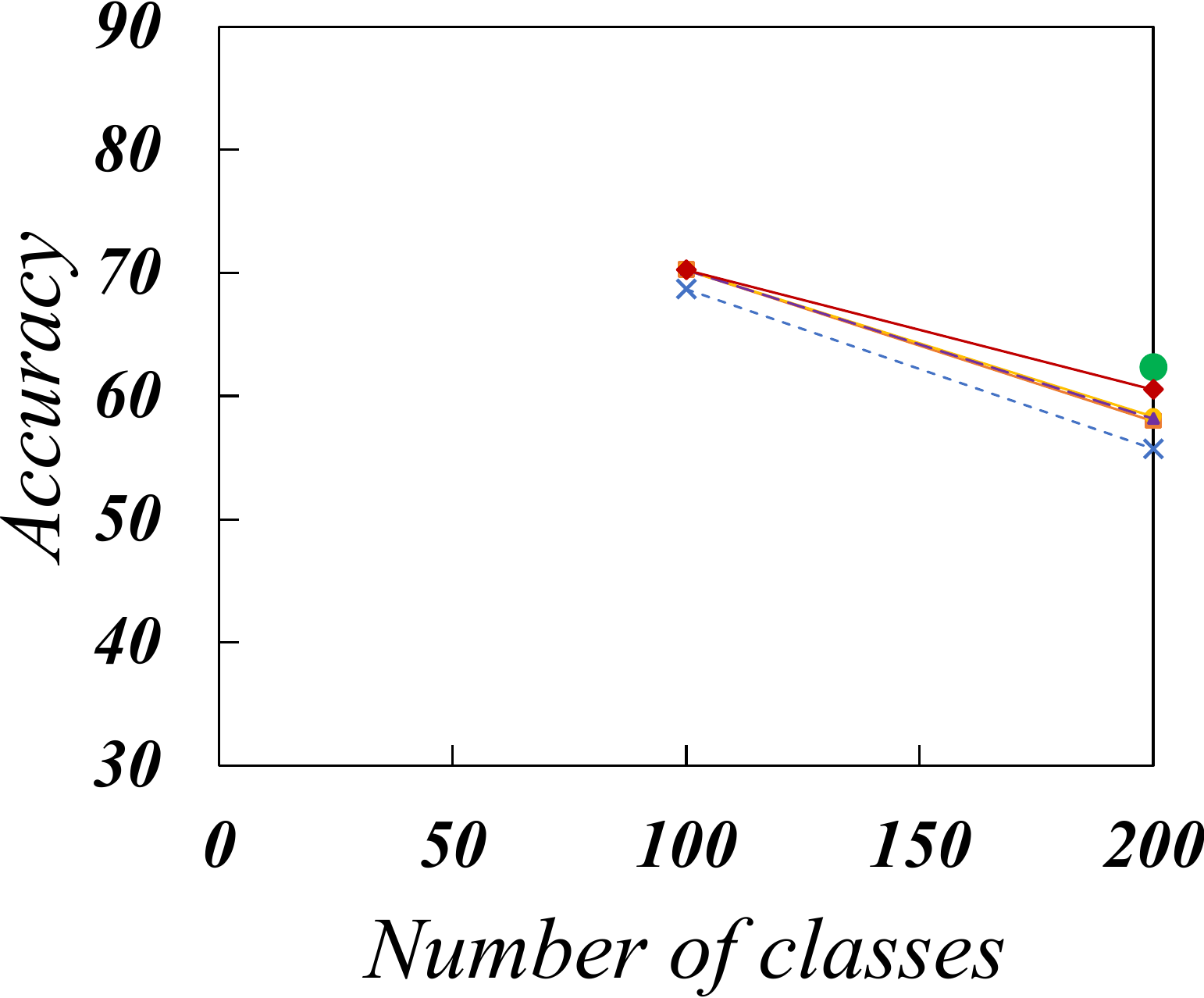}}  & 
			\subfigure[\label{fig:tiny:b}5 tasks] {\includegraphics[height=32mm]{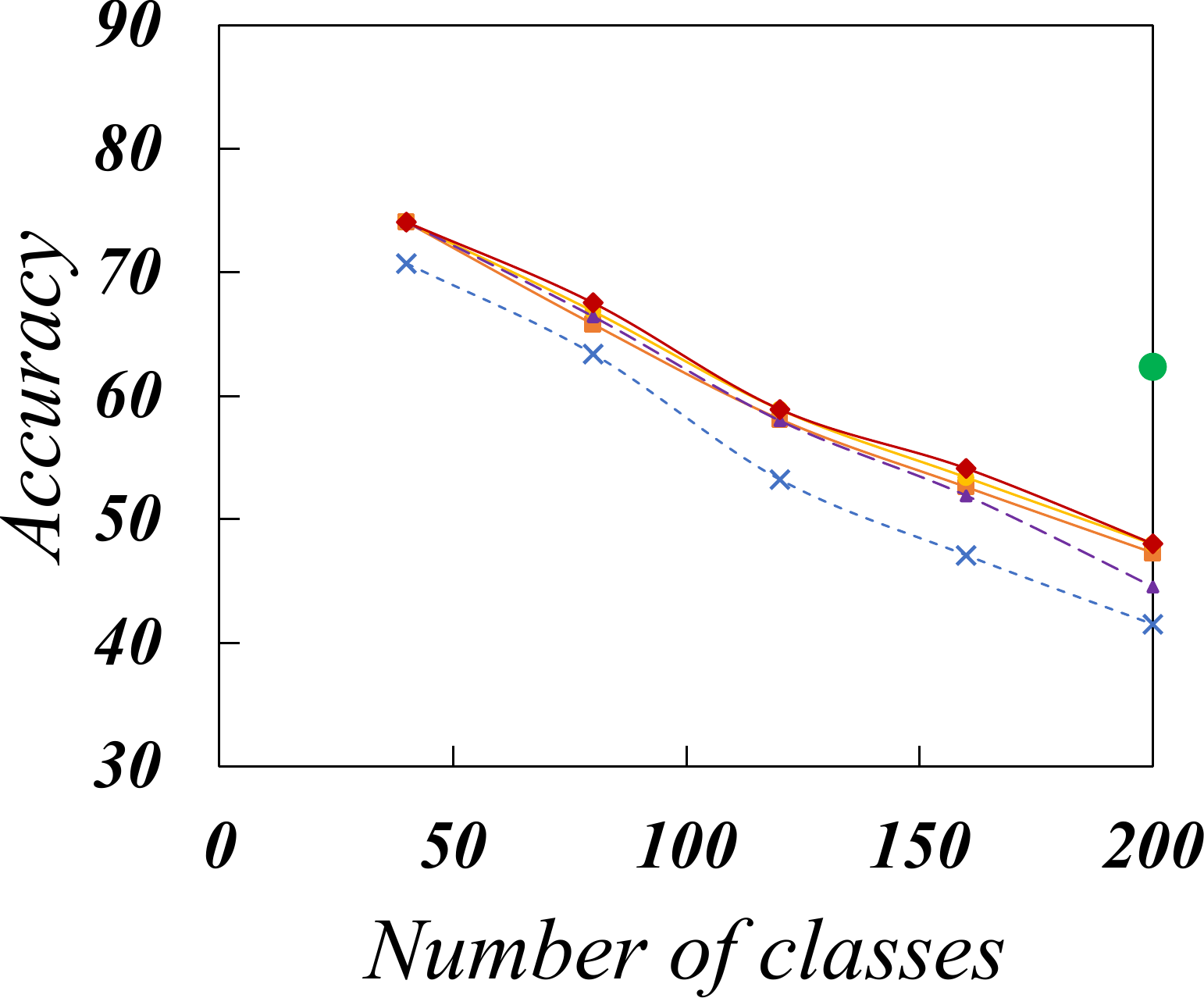}}   &
			\subfigure[\label{fig:tiny:c}10 tasks] {\includegraphics[height=32mm]{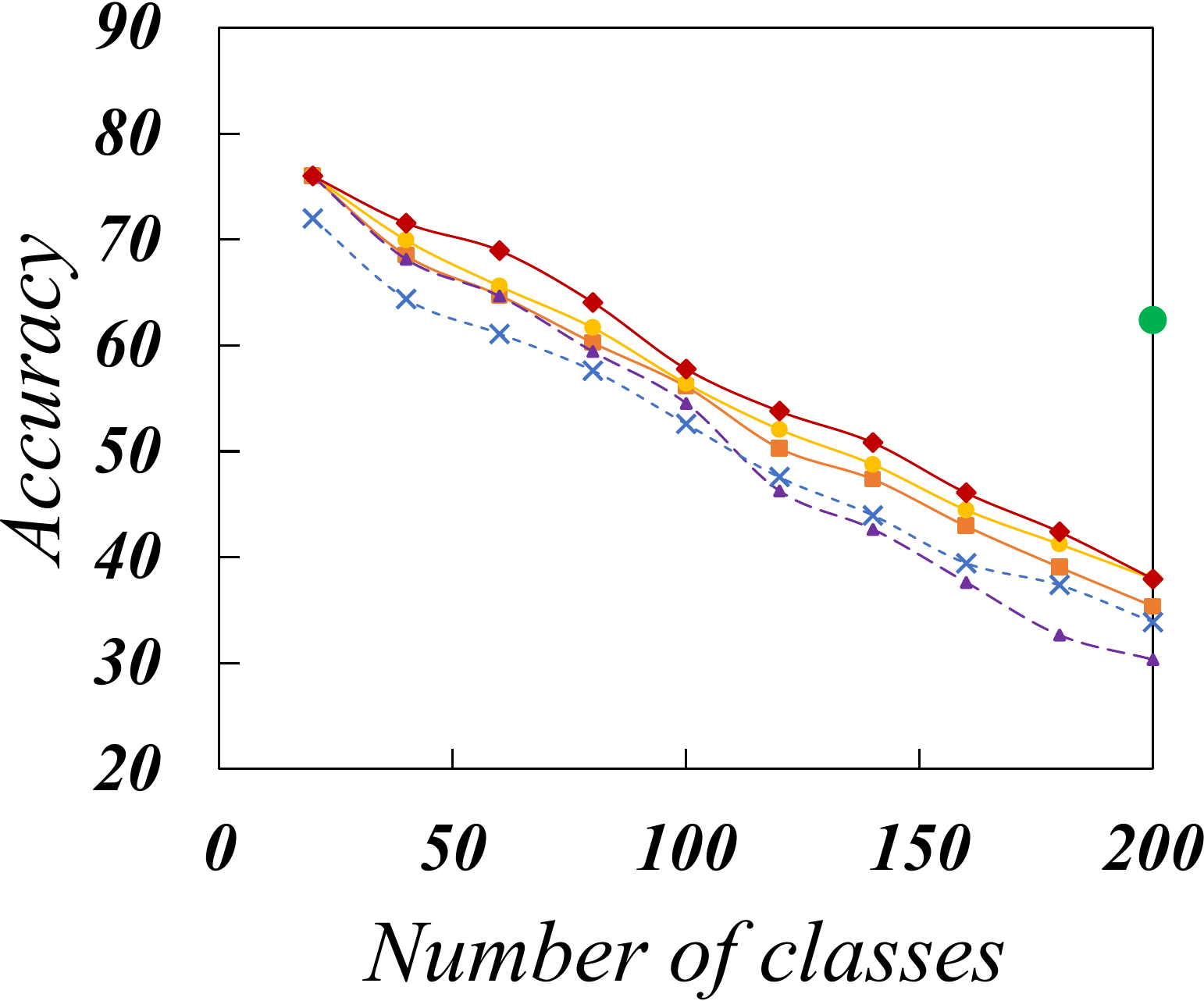}}  &
			\subfigure[\label{fig:tiny:d}20 tasks] {\includegraphics[height=32mm]{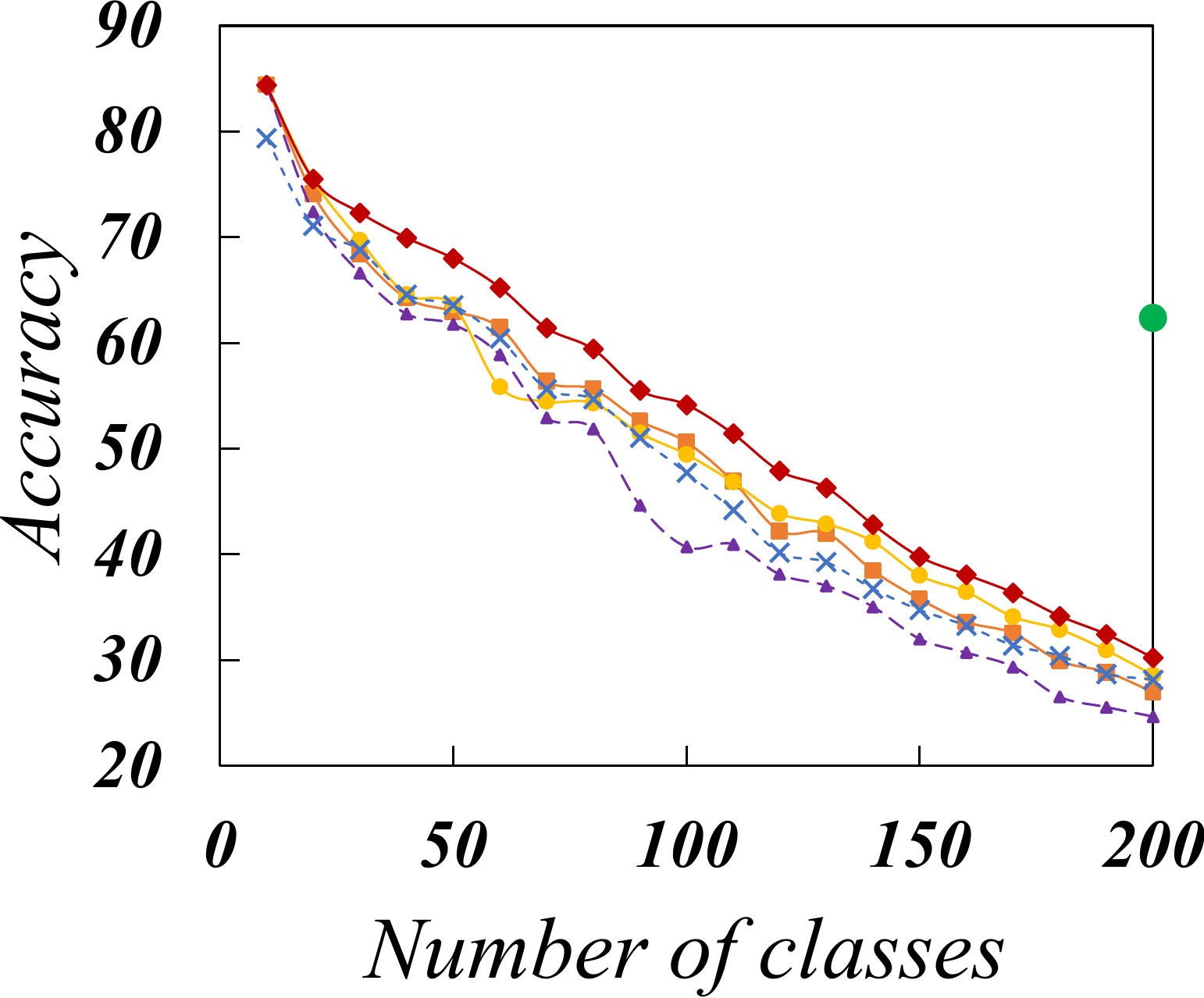}}  
		\end{tabular}
		\vspace{-3mm}
		\caption{Comparison on the accuracy for each incremental task using Tiny-ImageNet}
		\label{fig:tiny}
	}
	\vspace{3mm}
	\centering
	{
    \includegraphics[height=2.5mm]{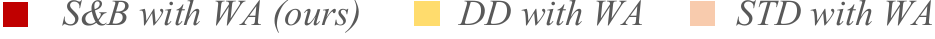}\vspace{1mm}

		\begin{tabular}{cccc}
			\subfigure[\label{fig:intra:a}Intra-new (CIFAR-100)] {\hspace{1mm}\includegraphics[height=34mm]{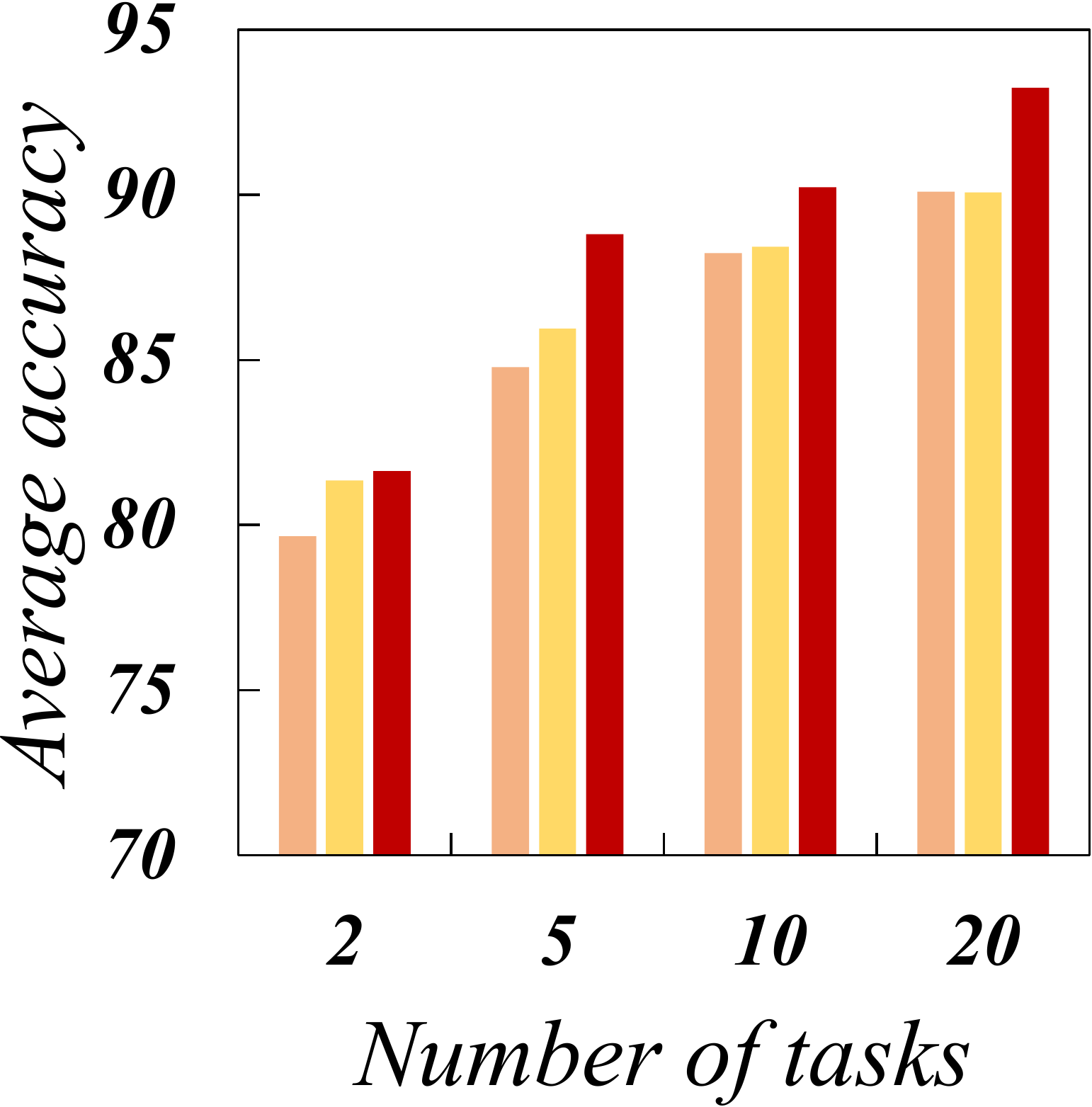}\hspace{3mm}}  & 
			\subfigure[\label{fig:intra:b}Intra-old (CIFAR-100)] {\hspace{1mm}\includegraphics[height=34mm]{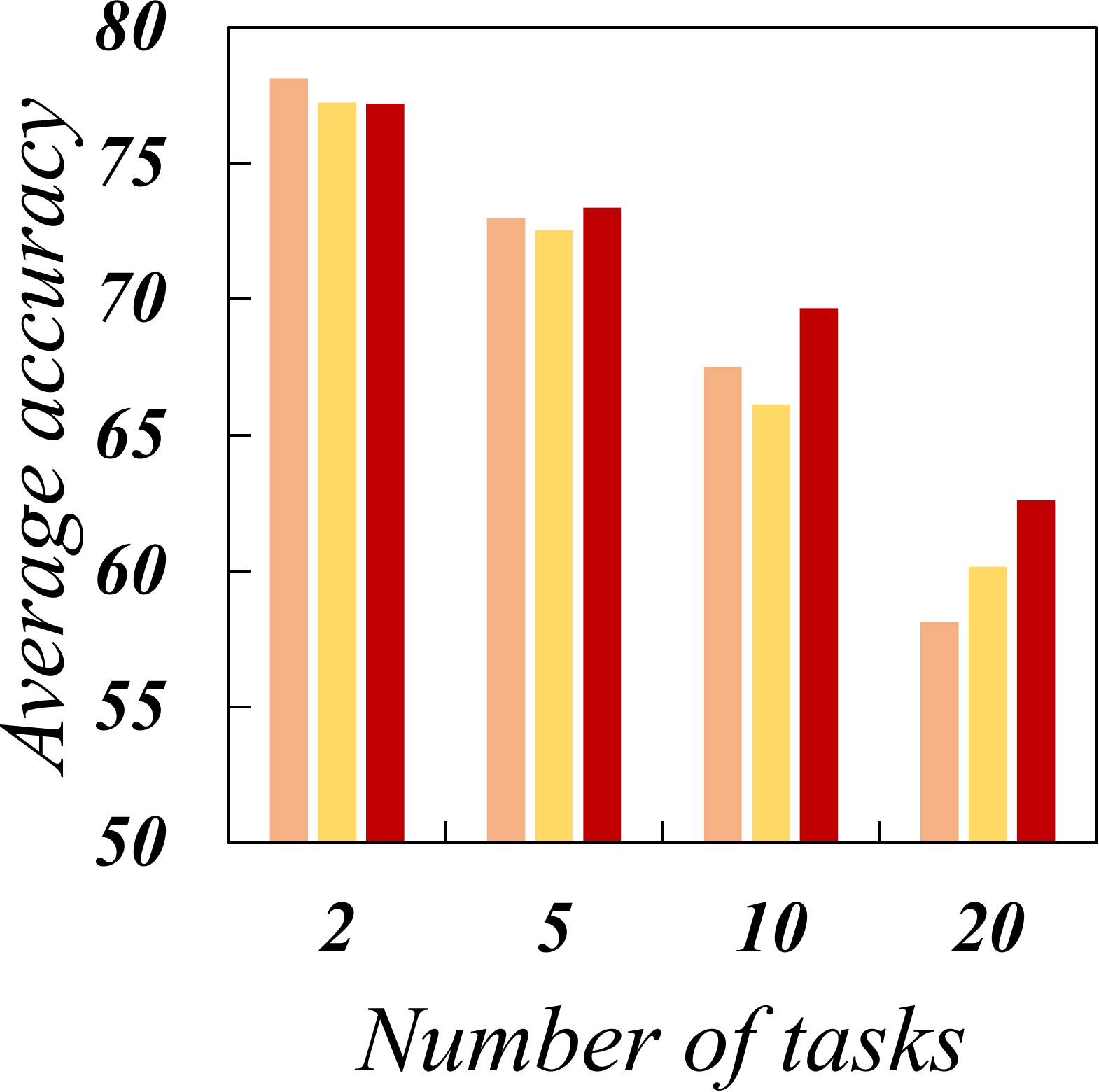}\hspace{3mm}}   &
			\subfigure[\label{fig:intra:c}Intra-new (Tiny-ImageNet)] {\hspace{1mm}\includegraphics[height=34mm]{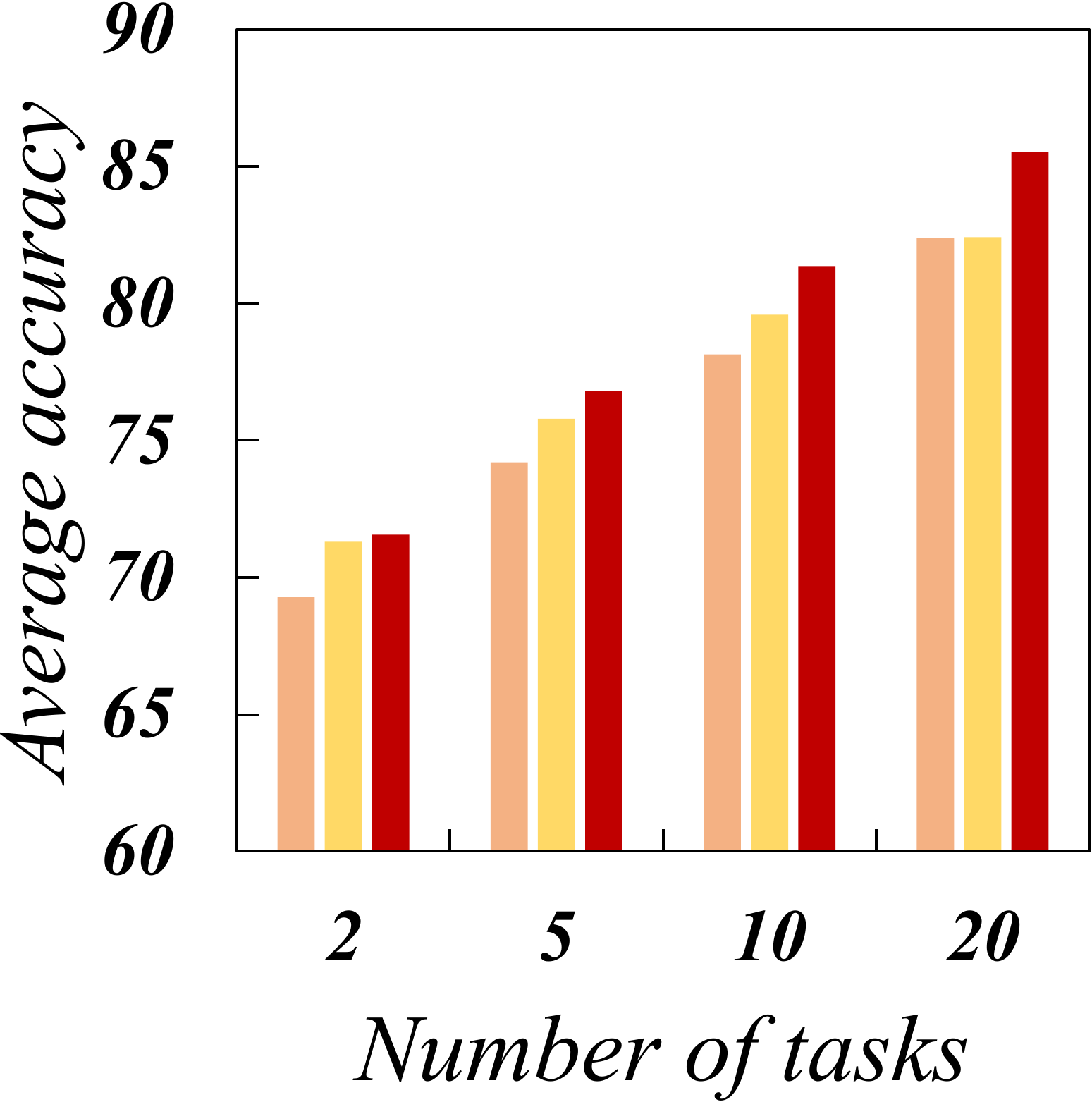}\hspace{4mm}}  &
			\subfigure[\label{fig:intra:d}Intra-old (Tiny-ImageNet)] {\hspace{1mm}\includegraphics[height=34mm]{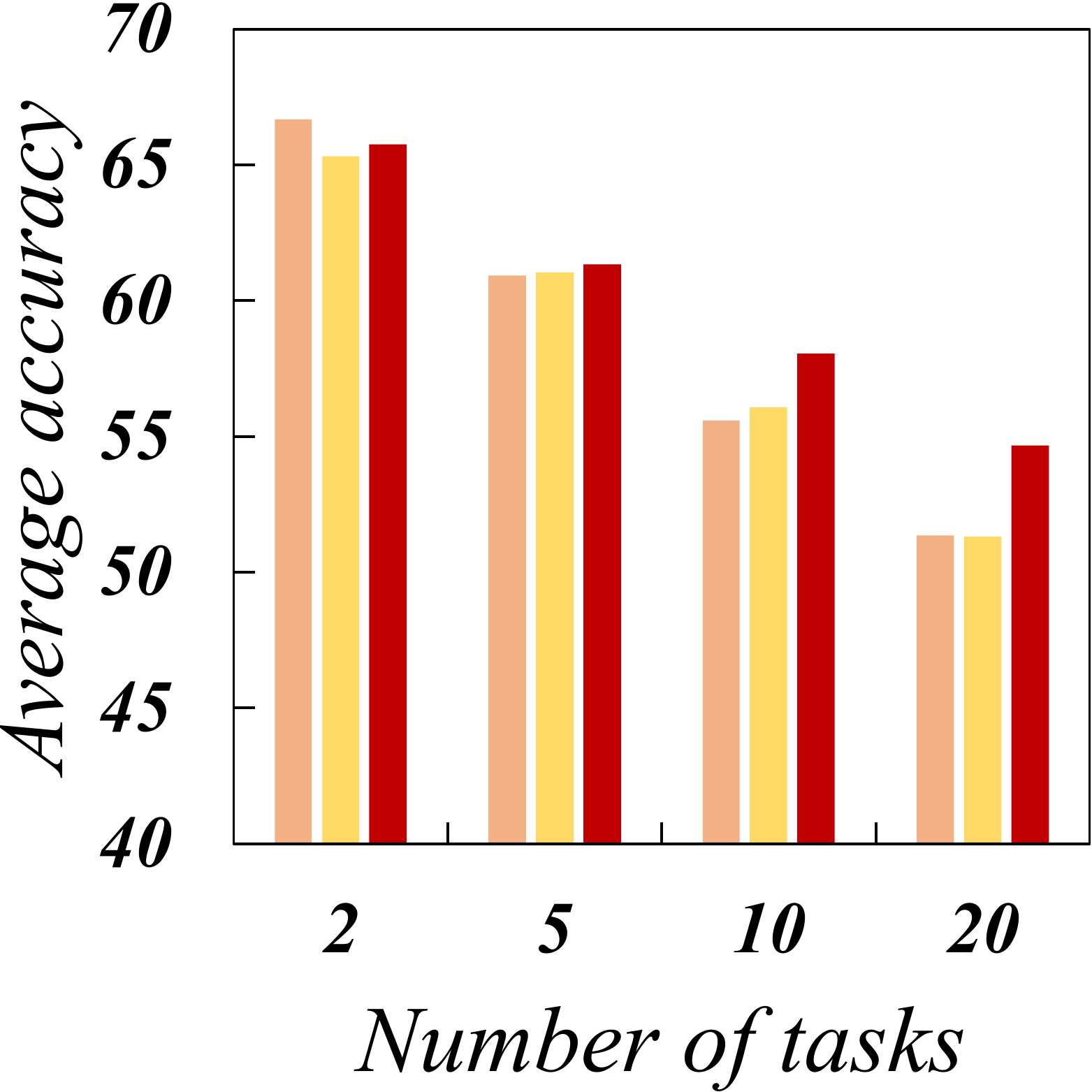}\hspace{3mm}}  
		\end{tabular}
		\vspace{-3mm}
		\caption{Comparison on the average intra-new and intra-old accuracy over the incremental tasks}
		\label{fig:intra}
	}
\end{figure*}

\subsection{Environment}
\smalltitle{Datasets and base model}
In our experiments, we train two benchmark datasets, CIFAR-100 \cite{krizhevsky2009learning} and Tiny-ImageNet \cite{le2015tiny}, on ResNet-18 \cite{HeZRS16}. CIFAR-100 consists of 50K training images, 500 per class, and 10K test images, 100 per class, from 100 classes in total, where every image is of size $32 \times 32$. We randomly arrange and divide the classes into a particular number of groups of the same size for each group to be an incremental task. Tiny-ImageNet includes 100K training images and 10K validation images for 200 classes, each of which is of size $64 \times 64$. As a labeled test set of Tiny-ImageNet is not available, we use its validation set as a test set. Similar to CIFAR-100, we randomly split the classes into a certain number groups of the same size to form a series of as many incremental tasks. As a base network, we use ResNet-18 for both datasets as it is widely used for various benchmark datasets including CIFAR-100 and Tiny-ImageNet in the literature \cite{abs-1708-04552,Keshari0V19}.

\smalltitle{Compared methods}
In order to evaluate the performance of Split-and-Bridge, we test the following KD-based class incremental learning methods: \textit{iCaRL} \cite{RebuffiKSL17}, \textit{WA} \cite{ZhaoXGZX20}, and \textit{Bic} \cite{WuCWYLGF19}. All three methods basically adopt the standard KD-based training method described in the preliminary section, which we name \textit{STD}\footnote{The original version of iCaRL is slightly different from STD, but this STD version is often reported to perform even better than the original one by the recent works such as \cite{DBLP:journals/corr/abs-2003-13947}.}. Their difference lies in their way of making inference. iCaRL makes a prediction by adopting $k$-NN classification on the exemplar set. Both WA and Bic, which are the state-of-the-art methods in CIL, propose new \textit{balancing} techniques to overcome the data imbalance problem between old classes and new classes, yet commonly follow STD for their training scheme. 

In addition to STD, we consider another baseline training method, called \textit{double distillation (DD)}, which is the way of combining two neural networks separately trained for either the old task and the new task as described in the motivation of our proposed method. Note that any inference techniques are orthogonal to the training scheme itself this work focuses on, and therefore we apply WA to DD as well as Split-and-Bridge. Finally, we measure the performance of \textit{oracle} as an upper bound, which is jointly trained at once using the entire dataset.


\smalltitle{Implementation details}
We implement all the methods\footnote{https://github.com/bigdata-inha/Split-and-Bridge} in PyTorch, and train each model on a machine with an NVIDIA TITAN RTX and Intel Core Xeon Gold 5122. In the split phase, we divide two last residual blocks along with the final fully-connected (FC) layer of ResNet-18 into two disjoint partitions, i.e., $\theta_o$ and $\theta_n$, implying $S=13$ and $L=18$. Full details are covered in our supplementary material. 


\subsection{Experimental Results}

\begin{table}[t]
    \centering
    \small
    \begin{tabular}{l||c|c|c|c}
          \hline
          \textbf{Number of tasks} & \textbf{2} & \textbf{5} & \textbf{10} & \textbf{20}  \\ 
         \hline\hline
        \multicolumn{5}{c}{CIFAR-100}  \\
          \hline
         STD with iCaRL  & 68.14 & 59.50 & 55.6 & 60.04  \\
         STD with Bic  & \textbf{69.96} & 67.07 & 60.65 & 49.89  \\
        STD with WA & 69.28 & 67.64 & 63.72 & 55.29 \\
       DD with WA & 68.84 & 67.68 & 63.12 & 58.08  \\
        S\&B with WA (ours) & 69.6 & \textbf{68.62} & \textbf{66.97} & \textbf{61.12} \\
        \hline
        Oracle  & \multicolumn{4}{c}{77.03} \\
        \hline
    \end{tabular}
    
    \begin{tabular}{l||c|c|c|c}
          \hline
         \multicolumn{5}{c}{Tiny-ImageNet}  \\
          \hline
         STD with iCaRL  & 55.72 & 51.32 & 48.65 & 46.56 \\
         STD with Bic  & 58.16 & 55.23 & 48.47 & 43.81 \\
        STD with WA & 57.96 & 55.97 & 51.61 & 47.57 \\
       DD with WA & 58.33 & 56.80 & 53.12 & 48.14 \\
        S\&B with WA (ours) & \textbf{60.52} & \textbf{57.16} & \textbf{54.81} & \textbf{51.63} \\
        \hline
        Oracle  &    \multicolumn{4}{c}{62.35} \\
        \hline
    \end{tabular}
    \caption{Average accuracies over all the incremental tasks of ResNet-18 using CIFAR-100 and Tiny-ImageNet}
    \label{tab:avgacc}
\end{table}

\smalltitle{Overall performance}
Table \ref{tab:avgacc} summarizes the overall performance of all the compared methods, where we present the average accuracy of each method over all the incremental steps other than the first. It is clearly observed that our Split-and-Bridge method outperforms the other learning methods in most of the results. As also shown in Figures \ref{fig:cifar} and \ref{fig:tiny}, Split-and-Bridge is not only the best on the average accuracy, but also consistently achieves the highest accuracy throughout the incremental steps in both CIFAR-100 and Tiny-ImageNet. The DD method occasionally performs better than STD probably because of learning the intra-new knowledge better, but its effectiveness turns out to be marginal in the sense that STD with WA or iCaRL often shows higher accuracies. This confirms our claim that a simple separated learning method with two neural networks does not work well in CIL.

Among the STD based methods, WA seems the best inference technique as it almost always shows the highest accuracy except for DD and Split-and-Bridge, and Bic is generally better than iCaRL in many cases. Interestingly, iCaRL performs quite well in the case of learning 20 tasks (i.e., 5 classes per task) in CIFAR-100, but both Bic and WA seem to struggle with learning this series of 20 tiny tasks as their performance gets abruptly worse than they used to be in the other cases. As observed in Figure \ref{fig:cifar:d}, this is due to the fact that the accuracy of the methods except for iCaRL and Split-and-Bridge is far lower than expected in the first couple steps, and these inaccurately trained initial models badly influences learning the next sequence of tasks. As mentioned earlier, starting with a high-capacity model can inevitably cause an overfitting problem in earlier steps especially when each task consists of a few classes such as one of 20 tasks in CIFAR-100. Since iCaRL does not involve any FC layer to make an inference, such an overfitting problem can be alleviated. Also, our Split-and-Bridge can mitigate overfitting with the help of regularization of the split phase, and hence still remains the best out of all the compared methods. For all the other details, please refer to our supplementary material.

\smalltitle{Accuracy on the intra-new and the intra-old knowledge}
As shown in Figure \ref{fig:intra}, we also compare the intra-new accuracy and the intra old accuracy of each method on the average. To focus on the effectiveness of each training scheme regardless of a balancing technique, we conduct this analysis using only the methods with WA. As expected, Split-and-Bridge always shows the best intra-new accuracy in both datasets, which tells our separated training scheme is quite effective to learn the intra-new knowledge. DD is also more adaptive to new tasks than STD when learning a small number of large tasks, but it gets less effective as the number of steps increases probably due to as many merging processes of two neural networks.

Another observation is that our Split-and-Bridge method is not only highly adaptive to new classes, but also good at preserving the intra-old knowledge as shown in Figures \ref{fig:intra:b} and \ref{fig:intra:d}. This is contrast to the result of Table \ref{tab:stdloss}, where learning by a CE loss without any KD losses is also adaptive but prone to forgetting the previous knowledge. Through this analysis, it is confirmed that Split-and-Bridge is able to achieve a good placement between \textit{stability} and \textit{plasticity}.

\section{Conclusion} \label{sec:con}
In class incremental learning, we need to learn three types of essential knowledge, intra-new, intra-old, and cross-task, but the standard KD-based method has more focused on preserving the intra-old knowledge by sacrificing plasticity of a model. Motivated by this, we proposed a novel class incremental learning method, called Split-and-Bridge, which is highly adaptive to new classes yet stable enough not to forget too much of the previous knowledge. Through the extensive experiments on CIFAR-100 and Tiny-ImageNet, we confirmed that our Split-and-Bridge method can be an effective solution for the \textit{stability-plasticity dilemma} in neural networks. As a future work, we hope to see our proposed training scheme is applied to a more complex deep learning problem such as object detection and sequence generation.

\section{Acknowledgements} \label{sec:ack}
This work was supported in part by the National Research Foundation of Korea (NRF) grant funded by the Korea government (MSIT) (NRF-2018R1D1A1B07049934) and in part by Institute of Information \& communications Technology Planning \& Evaluation (IITP) grants funded by the Korea government (MSIT) (2019-0-00240, 2019-0-00064, and 2020-0-01389, Artificial Intelligence Convergence Research Center(Inha University)).

\bibliography{sbbib}   

\end{document}